\documentclass[conference]{IEEEtran}
\IEEEoverridecommandlockouts
\usepackage[letterpaper, left=0.6in, right=0.6in, top=0.7in, bottom=1in]{geometry}

\usepackage[T1]{fontenc}
\usepackage{placeins}

\usepackage{graphicx}           
\usepackage{booktabs}           
\usepackage[ruled]{algorithm2e} 
\usepackage{cite}               
\usepackage{adjustbox}
\usepackage{enumitem}
\usepackage{amsmath}
\usepackage{multirow}
\usepackage{algorithmic}
\usepackage{amsfonts}
\usepackage{amssymb}
\usepackage[bookmarks=false]{hyperref}
\usepackage{comment}
\usepackage{xcolor}
\usepackage{CJK}
\def\BibTeX{{\rm B\kern-.05em{\sc i\kern-.025em b}\kern-.08em
    T\kern-.1667em\lower.7ex\hbox{E}\kern-.125emX}}

\usepackage{pifont} 
\usepackage{makecell} 

\newcommand{\cmark}{\ding{51}} 
\newcommand{\xmark}{\ding{55}} 

\begin{document}
\begin{CJK}{UTF8}{bkai}

\title{Watermarking Kolmogorov-Arnold Networks for Emerging Networked Applications via Activation Perturbation}

\author{\IEEEauthorblockN{Chia-Hsun Lu\textsuperscript{1}, Guan-Jhih Wu\textsuperscript{1}, Ya-Chi Ho\textsuperscript{2}, and Chih-Ya Shen\textsuperscript{1}}
\IEEEauthorblockA{\textsuperscript{1}\textit{Department of Computer Science}, \textit{National Tsing Hua University}, Hsinchu, Taiwan \\ \textsuperscript{2}\textit{Industrial Technology Research Institute}, Hsinchu, Taiwan \\
Email: chlu@m109.nthu.edu.tw, gjwu@lbstr.cs.nthu.edu.tw, light.ho@itri.org.tw, chihya@cs.nthu.edu.tw}}


%
%
%
\maketitle              

\begin{abstract}
With the increasing importance of protecting intellectual property in machine learning, watermarking techniques have gained significant attention. As advanced models are increasingly deployed in domains such as social network analysis, the need for robust model protection becomes even more critical. While existing watermarking methods have demonstrated effectiveness for conventional deep neural networks, they often fail to adapt to the novel architecture, Kolmogorov-Arnold Networks (KAN), which feature learnable activation functions. KAN holds strong potential for modeling complex relationships in network-structured data. However, their unique design also introduces new challenges for watermarking. Therefore, we propose a novel watermarking method, \emph{Discrete Cosine Transform-based Activation Watermarking (\texttt{DCT-AW})}, tailored for KAN. Leveraging the learnable activation functions of KAN, our method embeds watermarks by perturbing activation outputs using discrete cosine transform, ensuring compatibility with diverse tasks and achieving task independence. Experimental results demonstrate that \texttt{DCT-AW} has a small impact on model performance and provides superior robustness against various watermark removal attacks, including fine-tuning, pruning, and retraining after pruning.
\end{abstract}


\section{Introduction}

With the rapid development of machine learning, it has achieved significant success in various critical fields, such as recommendation systems and social network analysis. 
As the importance of deep learning models continues to grow, protecting their intellectual property becomes a challenge. To address this problem, model watermarking methods have gained interest. These methods involve embed identifiable features within the model, enabling effective ownership verification and enhancing the protection of intellectual property~\cite{li2019prove}. 

Generally, a good model watermarking technique should have the following two characteristics: i) \emph{Functionality preservation}: The embedding of the watermark should not significantly affect the model's performance. ii) \emph{Robustness against watermark removal attack}: The watermarking technique needs to resist watermark removal attacks and still be able to extract watermark information from the model. 
Traditional watermarking techniques have achieved some success in protecting in protecting deep neural networks (DNNs)~\cite{kim2023margin,chien2024customized}. However, these methods primarily focus on common neural network architectures (e.g. CNNs) and do not explore specialized models with unique and powerful representational capabilities. For instance, Kolmogorov-Arnold Networks (KAN)~\cite{liu2024kan} are a class of neural networks based on the Kolmogorov-Arnold representation theorem~\cite{kolmogorov1957representation}, have recently gained significant attention. With their exceptional expressive power, KAN is considered highly promising in various fields~\cite{genet2024tkan,cheon2024kolmogorov}. Even with such potential, intellectual property protection for KAN remains largely unexplored due to their relatively recent development.


An intuitive approach is to apply existing DNN watermarking techniques to KAN, but we find two main weaknesses. 

\noindent\textbf{Weakness W1. Task dependency of watermarking methods.} Many existing DNN watermarking techniques are designed with task-specific assumptions, limiting their generalizability to other problem settings. 

\noindent\textbf{Weakness W2. Vulnerability to attacks in KAN.} Due to the unique architecture and learnable activation functions of KAN, traditional DNN watermarking methods may not be as effective in watermark removal attacks such as retraining after pruning. 

To address these challenges, we propose \emph{\textbf{D}iscrete \textbf{C}osine \textbf{T}ransform-based \textbf{A}ctivation \textbf{W}atermarking (\texttt{DCT-AW})}, a novel watermarking method that embeds watermarks by perturbing activation functions in KAN. This approach effectively mitigates the above weaknesses by being task-independent and more resistant to attacks. 

We demonstrate the effectiveness of \texttt{DCT-AW} through comprehensive experiments compared with multiple state-of-the-art DNN watermarking approaches. The results show that our proposed method achieves strong performance in functional preservation 
and outperforms state-of-the-art methods against different watermark removal attacks, i.e., fine-tuning attacks, pruning attacks, and retraining after pruning attacks, when KAN is applied to different task, including classification and regression. 
Our contributions can be summarized as follows:
\begin{enumerate}[label=$\bullet$]
    \item We observe two major weaknesses, \textbf{W1} and \textbf{W2}, when applying existing watermarking approaches to Kolmogorov-Arnold Networks (KAN), and conduct a preliminary analysis to illustrate them.
    \item We propose \emph{Discrete Cosine Transform-based Activation Watermarking (\texttt{DCT-AW})}, the watermarking framework designed specifically for KAN. 
    \item Experimental results 
    show that our proposed \texttt{DCT-AW} are task-independent and more robust against various watermark removal attacks outperforming the state-of-the-art baselines.
\end{enumerate}


The rest of this paper is organized as follows. Sec.~\ref{c:related works} introduces the related works. 
Sec.~\ref{c:preliminary} provides the details about KAN 
and introduces the problem definition. 
Sec.~\ref{c:preliminary analysis} conducts the preliminary analysis. 
Sec.~\ref{c:proposed scheme} provides the details of our proposed method. Sec.~\ref{c:experiment} presents the experimental results. Finally, Sec.~\ref{c:conclusion} concludes this paper.

\section{Related Works}
\label{c:related works}
\noindent\textbf{DNN Watermark technique.} Watermarking has long been used to protect intellectual property. Uchida et al.~\cite{uchida2017embedding} propose the first method for embedding watermarks into DNN parameters. Since then, many methods have been developed. 
\emph{Trigger-set-based methods} use specific samples as key samples, which are embedded into the model as watermarks. These key samples can be used to verify the watermark in the model. Adi et al.~\cite{adi2018turning} and Zhang et al.~\cite{zhang2018protecting} both generate trigger samples with pre-defined labels as the watermarks and then train the watermarked model with the training set and trigger set. 
\emph{Feature-based methods} embed specific information into the model to incorporate the watermark. This method also requires providing a corresponding extractor to perform parameter extraction for ownership verification on suspicious models in the future. Fan et al.~\cite{fan2021deepipr} propose a passport-based watermarking method; this approach deteriorates DNN performance with forged passports, thus providing strong protection against forged watermarks. 
\emph{Signal-based methods} embed watermarks by adding perturbations to the model's output. The model owner can claim ownership by observing whether the suspicious model's output contains this specific perturbation. 
Chien et al.~\cite{chien2024customized} present customized soft-label perturbation to embed different perturbations into different models, achieving customized watermarking effects and resisting existing watermark removal attacks.

\noindent\textbf{Kolmogorov-Arnold Networks (KAN).} Kolmogorov-Arnold Networks (KAN)~\cite{liu2024kan} represent a novel approach in the landscape of neural network architectures. Traditional multi-layer perceptrons (MLP) is designed based on the universal approximation theorem~\cite{hornik1989multilayer}, whereas KAN is inspired by the Kolmogorov-Arnold representation theorem~\cite{arnol1959representation, kolmogorov1961representation}. 
Many techniques have been developed based on KAN. Abueidda et al.~\cite{abueidda2024deepokan} replace B-spline functions with Gaussian radial basis functions in KAN neurons, achieving significant speedup and applying the method to various mechanical problems. 

Considering the development and potential of KAN, there is a need to propose a method for protecting the intellectual property of KAN model architectures. However, no watermarking methods specifically designed for KAN have been proposed so far, highlighting the need for such an approach. 

\section{Preliminary}
\label{c:preliminary}


\subsection{Details of Kolmogorov-Arnold Networks (KAN)}
Kolmogorov-Arnold Networks (KAN)~\cite{liu2024kan} replace fixed activation functions in traditional networks (e.g., MLPs) with learnable spline-based functions. The design is inspired by the Kolmogorov-Arnold theorem, which states that any continuous multivariate function can be expressed as a sum of univariate functions. KAN improves expressiveness by learning activation behaviors.
Generally, KAN can be expressed as follow:
\begin{equation}
\text{KAN}(x) = (\Psi_{L-1} \circ \Psi_{L-2} \circ \cdots \circ \Psi_1 \circ \Psi_0)x.
\end{equation}

Where $\Psi_l$ is the function matrix corresponding to the $l$-th KAN layer. Specifically, the activation function $\phi_{l,i,j} \in \Psi_l$ is a crucial component for the network's expressiveness and performance. In KAN, the activation function is constructed using B-splines, which are piecewise polynomial functions that provide a high degree of flexibility and control. The activation function $\phi(x)$ is composed of a basis function $b(x)$ and a spline function $spline(x)$:

\vspace{-0.8em}
\begin{equation} \label{fun:KAN_learn}
\phi(x) = w_b b(x) + w_s spline(x).
\end{equation}
\vspace{-1em}

The basis function $b(x)$ is typically chosen as the SiLU (Sigmoid Linear Unit, silu) activation function and the spline function $spline(x)$ is parameterized as a linear combination of B-splines. 
The weights $w_b$ and $w_s$ control their respective contributions. This architecture enables KAN to flexibly learn activation, improving its ability to represent complex patterns.


\subsection{Problem Formulation}
\label{c:problem definition}

\subsubsection{Machine Learning Model Watermarking}
\begin{figure}[h]
    \centering
    \includegraphics[width=0.8\columnwidth]{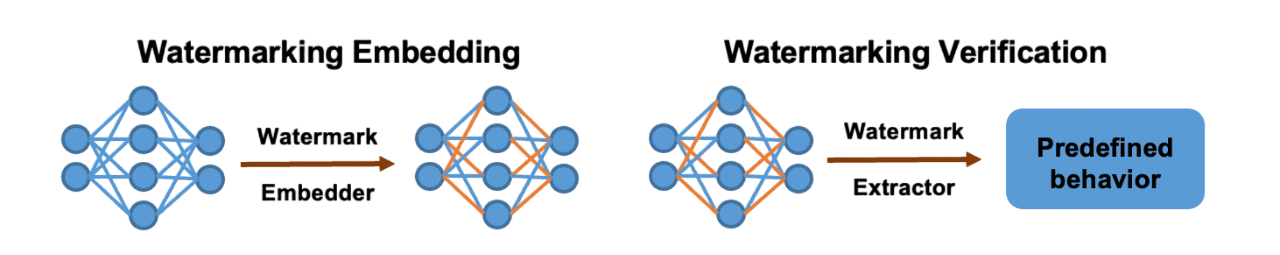} 
    \caption{Illustration of watermark embedding and verification process. In the watermark embedding step, specific information is embedded into the model. During the watermark verification step, this information is extracted in a specific way, and the model is determined to have the owner's embedded watermark if the extracted information matches the predefined behavior.}
    \label{fig:watermark_process}
\end{figure}

The machine learning model watermarking framework typically consists of two steps: 1) watermark embedding and 2) watermark verification. Figure~\ref{fig:watermark_process} illustrates the framework of the watermark embedding and watermark verification step.

\noindent\textbf{Watermark embedding.} Given a machine learning model $M_m$ and watermark information $w$ that to be embedded. The embedding process can be expressed as:
    \begin{equation}
        M_{\text{wm}} = \text{Embed}(M_m, w),
    \end{equation}
    where $M_{\text{wm}}$ is the resulting model with the watermark.

\noindent\textbf{Watermark verification.} Let $M_v$ denote the model subjected to verification, and $V_{\text{wm}}$ represents the outcome of the verification process. In this step, we assess how effectively the embedded information $w$ can be retrieved from $M_v$. The verification process is described as:
    \begin{equation}
        V_{\text{wm}} = \text{Verify}(M_v, w).
    \end{equation}
    A higher value of $V_{\text{wm}}$ indicates a stronger detection of the embedded watermark $w$.

\vspace{-1em}
In the following, we briefly introduce the common \emph{watermark removal attacks}. 
\textbf{Fine-tuning attack} is a technique where a pre-trained model is further trained on a new dataset or the same dataset. When applied as a watermark removal attack, this technique can alter the distribution of the model and potentially remove the watermark information during the training process. 
\textbf{Pruning attack} 
affects watermarking methods that embed watermark information into the model's weights. In such cases, the watermark information may be set to zero during the pruning process, leading to the removal of the watermark. 
\textbf{Retrain after pruning attack} builds upon the pruning attack. While pruning alone can disrupt watermark information, it may also degrade the model's original performance if critical weights are removed. To recover model utility, attackers may retrain the pruned model, aiming to restore accuracy while maintaining the watermark removal. This two-step process poses a serious threat to watermark robustness.

After outlining the watermark embedding process, verification strategies, and common watermark removal attacks, we now formally define the objective of our task.

\subsubsection{Problem Definition}
This paper aims to design a machine learning model watermarking framework that maximizes detection accuracy while satisfying the following requirements:

\noindent\textbf{Functionality-Preservation.} The watermark embedding process should have a small impact on the performance of the main task.
    

    
    
\noindent\textbf{Dependency with Different Task (addressing Weakness W1).} The proposed framework is independent of the task, i.e., the proposed watermark framework can apply to different tasks.
    
    
\noindent\textbf{Robustness against Removal Attacks (addressing Weakness W2).} The watermark should remain exist within the model after it suffers from the watermark removal attacks, such as fine-tuning attacks, pruning attacks, and retrain after pruning attacks.
    

\section{Preliminary Analysis}
\label{c:preliminary analysis}

In this section, we discuss two major weaknesses of applying existing DNN watermarking techniques to KAN and highlight the motivation for developing a tailored watermarking approach for KAN.

\subsection{Weakness W1: Task Dependency of Watermarking Methods}

Existing DNN watermarking methods typically fall into two categories: architecture-related (e.g., feature-based) and task-related (e.g., trigger-set-based and signal-based). However, both types face limitations when applied to KAN due to fundamental architectural differences. Architecture-based methods often embed watermarks into model weights, which is incompatible with KAN's design that relies on learnable activation functions instead of fixed ones. Task-based methods modify decision boundaries or embed specific output signals, making them suitable for classification tasks but difficult to extend to regression. Thus, their scalability across tasks remains limited. 

\subsection{Weakness W2: Vulnerability to Attacks in KAN}
\label{sec:discussion}



Watermark removal attacks pose significant threats to the integrity and security of neural networks. To evaluate the robustness of existing DNN watermarking methods when applied to KAN, we assess four representative techniques: \texttt{WNN}~\cite{adi2018turning}, \texttt{Margin}~\cite{kim2023margin}, \texttt{Blind}~\cite{li2019prove}, and \texttt{USP}~\cite{chien2024customized}, under various types of watermark removal attacks. The results are summarized in Table~\ref{table:methodstoattacks}.

\vspace{-1.2em}
\begin{table}[ht]
\centering
\caption{Robustness of neural network watermarking methods to different attacks on KAN. The mark \cmark{} means the watermarking method is robust to the attack. The mark \xmark{} means the watermarking method cannot resist the attack.}
\label{table:methodstoattacks}
\resizebox{0.8\columnwidth}{!}{
\def\arraystretch{1.2}
\begin{tabular}{ |c||c|c|c|c| } 
\hline
\textbf{Method} & \makecell{\Gape[2pt] {\textbf{Fine-tuning}}\\\textbf{small lr}} & \makecell{\Gape[2pt] {\textbf{Fine-tuning}}\\\textbf{large lr}} & \textbf{Pruning} & \textbf{Re-training} \\
\hline
\texttt{WNN}~\cite{adi2018turning} (USENIX' 2018) & \cmark & \cmark & \xmark & \xmark \\ 
\texttt{Margin}~\cite{kim2023margin} (ICML' 2023) & \cmark & \xmark & \cmark & \xmark \\ 
\texttt{Blind}~\cite{li2019prove} (ACSAC' 2019) & \cmark & \cmark & \xmark & \xmark \\ 
\texttt{USP}~\cite{chien2024customized} (WSDM' 2024) & \cmark & \cmark & \cmark & \xmark \\ 
\hline
\end{tabular}
}
\end{table}
\vspace{-1em}

The results show that most existing methods fail to defend against pruning or retraining attacks on KAN, which demonstrates the insufficiency of current watermarking strategies in this setting. To understand the underlying cause of this vulnerability, we analyze the differences between MLP and KAN that contribute to this weakness and discuss why applying DNN watermarking techniques directly to KAN may be ineffective.

To justify vulnerability to attacks in KAN, we conduct experiments comparing MLP and KAN under attack. Here, we use pruning attack. Specifically, MLP and KAN architectures with identical structures. 
We utilize the MNIST dataset and trained both models for 10 epochs. The pruning rate of pruning attack is incrementally applied from 0\% to 100\% in steps of 10\%. The objective was to observe the changes in accuracy and loss after pruning.


\begin{table}[!t]
\caption{Pruning results for MLP and KAN models.}
\label{table:pruning_results}
\centering
\resizebox{0.7\columnwidth}{!}{
\def\arraystretch{1.2}
\begin{tabular}{|c|cc|cc|}
\hline
\multirow{2}{*}{\textbf{Pruning Ratio(\%)}} & \multicolumn{2}{c|}{\textbf{MLP}} & \multicolumn{2}{c|}{\textbf{KAN}} \\ \cline{2-5} 
 & \textbf{Loss} & \textbf{Acc} & \textbf{Loss} & \textbf{Acc} \\ \hline
0 & 0.1031 & 97.06\% & 0.1125 & 96.68\% \\ \hline
10 & 0.1031 & 97.08\% & 2.1639 & 29.06\% \\ \hline
20 & 0.1032 & 97.00\% & 2.7977 & 12.59\% \\ \hline
30 & 0.1031 & 97.04\% & 2.7077 & 9.74\% \\ \hline
40 & 0.1064 & 97.03\% & 2.3688 & 8.88\% \\ \hline
50 & 0.1048 & 96.85\% & 2.3000 & 8.91\% \\ \hline
60 & 0.1123 & 96.58\% & 2.2997 & 6.99\% \\ \hline
70 & 0.2264 & 93.84\% & 2.3028 & 7.48\% \\ \hline
80 & 1.1356 & 82.28\% & 2.3023 & 4.37\% \\ \hline
90 & 1.4046 & 78.05\% & 2.3026 & 10.32\% \\ \hline
100 & 2.3082 & 9.74\% & 2.3026 & 9.80\% \\ \hline
\end{tabular}
}
\end{table}


Table \ref{table:pruning_results} shows that pruning has less impact on MLP models. 
The loss does not significantly increase until 70\% of the weights are pruned. In this scenario, retraining does not cause substantial modifications to the model, thus preserving the watermark information. In contrast, pruning has a more pronounced effect on KAN. Even at lower levels of pruning, the accuracy drops significantly, and the loss increases sharply. It demonstrates that the watermark information in KAN is more susceptible to being disrupted by retrain after pruning attacks. These findings show that existing DNN watermarking methods, which rely on the stability of weights and redundancy in MLP, are less effective when applied to KAN. The distinct behavior of KAN under pruning highlights the need for developing watermarking techniques specifically tailored to their unique architecture.


\section{Proposed Scheme}
\label{c:proposed scheme}

In order to address the two weaknesses \textbf{W1} and \textbf{W2} outlined in Sec.~\ref{c:preliminary analysis}, we propose \emph{\textbf{D}iscrete \textbf{C}osine \textbf{T}ransform-based \textbf{A}ctivation \textbf{W}atermarking (\texttt{DCT-AW})}, a novel scheme tailored for KAN. In this section, we first present the watermark embedding process using frequency-domain perturbations, and then introduce the detector training and verification process.


\subsection{Watermark Embedding Process of DCT-AW}

    
\vspace{-1.2em}
\begin{figure}[!ht]
    \begin{minipage}{0.48\linewidth}
    \centering
    \includegraphics[width=0.7\textwidth]{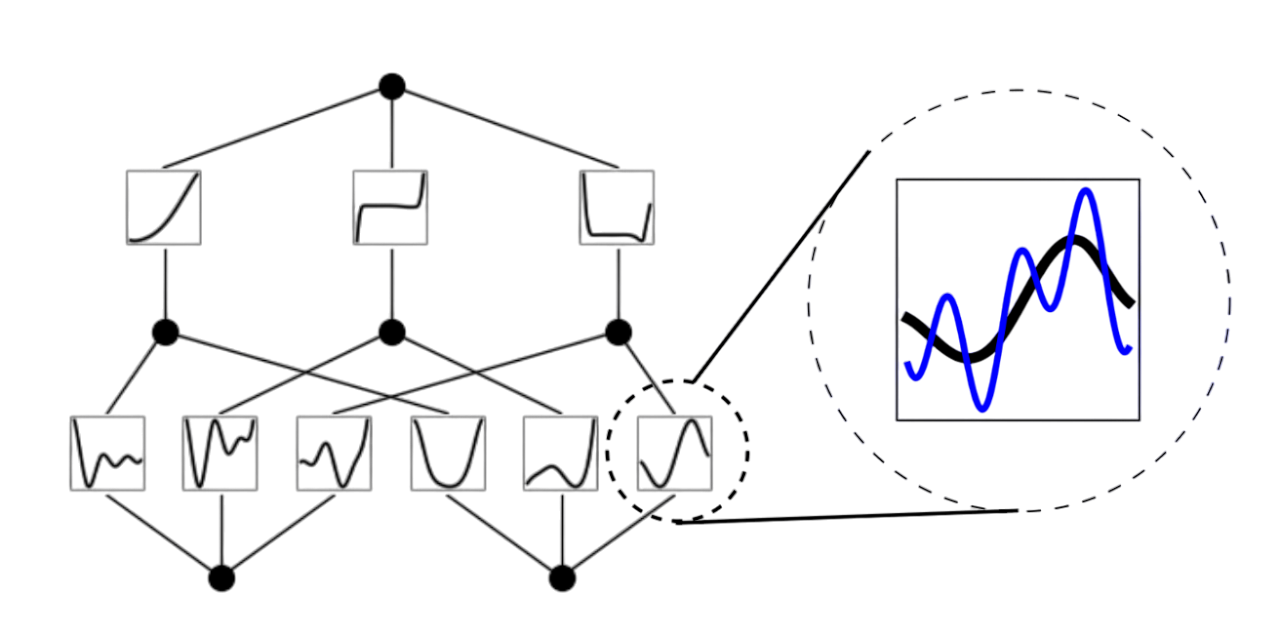}
    \caption{Concept of the proposed approach \texttt{DCT-AW}.} 
    \label{fig:concept}
    \end{minipage}
    \begin{minipage}{0.48\linewidth}
    \centering
    \includegraphics[width=\textwidth]{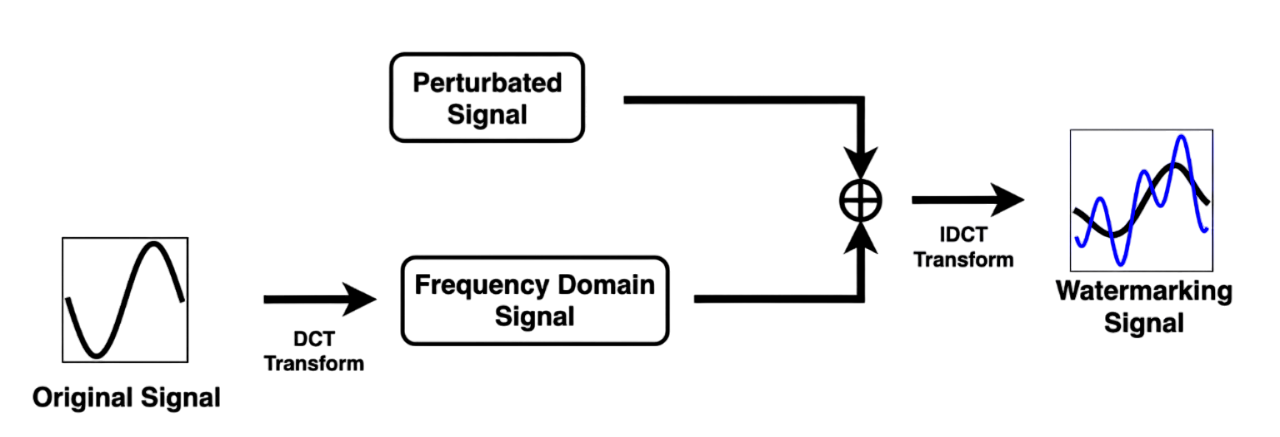}
    \caption{Signal generation framework using DCT for watermark embedding.} \label{fig:framework}
    \end{minipage}
\end{figure}
\vspace{-0.8em}

The basic idea of \texttt{DCT-AW} is to embed watermarks into the learnable activation functions of KAN by introducing subtle perturbations in the frequency domain. Figure \ref{fig:concept} shows the concept of \texttt{DCT-AW}. During training, we introduce our perturbations into the model to ensure that specific activation functions exhibit unique signals different from those in a clean model, while maintaining the original performance of the model. During the verification step, the presence of these perturbations is detected to confirm the embedded watermark. 


To address weakness \textbf{W1}, which concerns the task dependency of traditional watermarking methods, \texttt{DCT-AW} adopts an architecture-based embedding strategy. Specifically, it leverages the architecture of KAN, in which \emph{activation functions} are learnable, allowing us to embed perturbations directly into selected activation functions. Unlike trigger-set-based approaches that require task-specific data, \texttt{DCT-AW} embeds the watermark into the model architecture itself. This design enables task-agnostic watermarking and ensures compatibility across different tasks.


To address weakness \textbf{W2}, which concerns KAN's vulnerability to pruning and retraining, \texttt{DCT-AW} introduces a frequency-domain perturbation strategy. We apply the Discrete Cosine Transform (DCT)~\cite{ahmed1974discrete} to activation outputs, embed the watermark in the frequency domain, and reconstruct the perturbed outputs using the inverse DCT (IDCT). This allows fine-grained adjustments while preserving the overall structure of the activation functions. 
Formally, DCT can be expressed as follows: $X_k = \sum_{n=0}^{N-1} x_n \cos \left[ \frac{\pi}{N} \left( n + \frac{1}{2} \right) k \right]$, 
where $ x_n $ is the activation function output and $ k $ is the frequency index. Similarly, IDCT can be expressed as follows: $x_n = \frac{1}{N} \sum_{k=0}^{N-1} X_k \cos \left[ \frac{\pi}{N} \left( n + \frac{1}{2} \right) k \right]$, 
where $ X_k $ is the DCT-transformed signal and $ n $ is the sample index. In \texttt{DCT-AW}, we perturb the frequency components via:
\begin{equation}
y' = \text{IDCT}(\text{DCT}(y) + P)
\end{equation}
where $y$ represents the original layer $0$ activation function outputs, $P$ is the perturbation signal, and $y'$ is the perturbed activation function output.

Figure~\ref{fig:framework} illustrates the generation process of watermarked signal. By using the DCT and Inverse DCT techniques, we can generate suitable signals for the activation function. This allows the model to learn distinct characteristics that differentiate it from the clean model. 

After generating the watermark signal, we investigate where to embed it in the model. Since KAN has multiple learnable activation layers, we compare performance when embedding into layer $0$ and layer $1$, using both a classification task (MNIST~\cite{deng2012mnist}) and a regression task (Feynman dataset~\cite{udrescu2020ai,udrescu2020ai2}). 
Table~\ref{tab:layer_comparison} shows that embedding the watermark into layer $0$ leads to less performance degradation than into layer $1$, for both classification and regression. This is because earlier layers have weaker influence on the final output, making them more suitable for subtle modifications. In contrast, deeper layers affect predictions more directly, amplifying the watermark's impact. Therefore, we embed the watermark into layer $0$ in \texttt{DCT-AW} to balance detectability and task performance.

\begin{table}[!t]
    \caption{Performance with watermark embedded in different layers. $\uparrow$ indicates higher is better; $\downarrow$ indicates lower is better.}
    \label{tab:layer_comparison}
    \centering
    \begin{adjustbox}{width=0.6\columnwidth}
    \def\arraystretch{1.1}
    \begin{tabular}{l|c|c}
        \hline
        \makecell{\Gape[2pt] Downstream \\ Task} & \makecell{\Gape[2pt] Watermark in \\ Layer $0$} & \makecell{\Gape[2pt] Watermark in \\ Layer $1$} \\
        \hline
        \makecell{\Gape[2pt] Classification \\ (Accuracy $\uparrow$)} & 95.09 & 88.74 (-6.35) \\
        \hline
        \makecell{\Gape[2pt] Regression \\ (RMSE $\downarrow$)} & 0.0397 & 0.1828 (+0.1431) \\
        \hline
    \end{tabular}
    \end{adjustbox}
\end{table}

After selecting the embedding location and designing a suitable watermark signal, the next step is to train the model such that the designated activation functions learn to incorporate the signal. 
The training process is outlined in Algorithm~\ref{alg1}. 

\begin{algorithm}
\caption{\small Embed watermark in KAN-based model}
\label{alg1}
\scriptsize
\begin{algorithmic}[1]
\REQUIRE Model $M_m$, Perturbation signal $w$, Number of Epochs $T$
\ENSURE Watermarked Model $M_{\text{wm}}$
\STATE Initialize the model parameters
\FOR{epoch = 1 to $T$}
\FOR{each batch of input data $\mathbf{x}$}
\STATE \textbf{Step 1:Normal Training Phase}
\STATE Perform a forward pass and compute the main task loss 
\STATE Perform backpropagation and update the model parameters
\STATE \textbf{Step 2:Activation Neuron Update Phase}
\STATE Perform a forward pass through layer $0$ to obtain outputs $O$ 
\STATE Compute the signal $O_{s}$ using $O_{s} = \mathrm{DCT}(O, w)$
\STATE \hspace{-0.3em}Calculate the signal loss $\mathcal{L}_{\text{Signal}} = \mathrm{MSE}(O, O_{s})$
\STATE \hspace{-0.3em}Perform backpropagation and update the parameters of layer $0$
\ENDFOR
\ENDFOR
\RETURN Watermarked Model $M_{\text{wm}}$
\end{algorithmic}
\end{algorithm}

In the first phase of Algorithm \ref{alg1}, we enhance the performance of the model on its original task. This phase's loss function and parameter updates are consistent with those of a clean model. In the second phase of Algorithm \ref{alg1}, we update the parameters of the activation functions of layer $0$ to embed the specific signal. We calculate the signal from the outputs of the activation functions of layer $0$ and then compute the signal loss. By updating the parameters of the activation functions of layer $0$ based on the signal loss $\mathcal{L}_{\text{Signal}}$, the activation function of the activation functions of layer $0$ learns the specific signal. 


\subsection{Detector Training and Watermark Verification Process of DCT-AW}
To detect the watermark, we train a three-layer MLP detector $M_d$ using binary classification. The input to the detector is the layer-$0$ activation output from both the watermarked model $M_{\text{wm}}$ and the clean model $M_m$, with labels $1$ and $0$, respectively. For each sample $d \in D$, we obtain $o^d_{\text{wm}}$ and $o^d_m$ as activation outputs and construct the training dataset accordingly.

To enhance generalization, we apply data augmentation by shuffling the order of activation values. 
Specifically, we randomizes the order of $o^d_{\text{wm}}$ and $o^d_m$. The resulting shuffled outputs are denoted as $o^d_{\text{wm}'}$ and $o^d_{m'}$, respectively. 
We generate 10 such variants for each $o^d_{\text{wm}}$ and $o^d_m$, and assign the same labels as their corresponding originals.


\vspace{-0.5em}
\begin{equation}\label{detector:dataset}
\text{Label}(o^d_n) = 
\begin{cases} 
1 & \text{if } n = {\text{wm}}\ \text{or}\ n = {\text{wm}}', \\
0 & \text{if } n = m\ \text{or}\ n = m'.
\end{cases}
\end{equation}
\vspace{-0.5em}

Let $O_{\text{wm}} = \{o^d_{\text{wm}} | d \in D\}$, $O_{\text{wm}'} = \{o^d_{\text{wm}'} | d \in D\}$, $O_m = \{o^d_m | d \in D\}$, and $O_{m'} = \{o^d_{m'} | d \in D\}$. We construct the detector training dataset $D_{det} = (X_{det}, Y_{det})$, where $X_{det} = \{O_{\text{wm}} \cup O_{\text{wm}'}, O_m \cup O_{m'}\}$ represents the inputs and $Y_{det} = \{1, 0\}$ contains the labels assigned as defined in Eq.~\ref{detector:dataset}. The detector is then trained using binary classification. For this task, we utilize cross-entropy to compute the detection loss $\mathcal{L}_{\text{Detect}}$, defined as:

\vspace{-0.8em}
\begin{equation}
\mathcal{L}_{\text{Detect}} = \text{CrossEntropy}(M_d(X_d), Y_d).
\end{equation}
\vspace{-1em}

To verify whether a suspect model $M_s$ has been watermarked, the suspect model $M_s$ is tested on a separate testing dataset that has not been seen by either $M_s$ or the detector. The outputs of $M_s$ are then fed into the detector model $M_d$. The detector analyzes the output distribution to determine whether $M_s$ contains the watermark.

\begin{table*}[!h]
    \centering
    \caption{Performance comparison on MNIST.}\label{table:dif_attack_mnist}
    \begin{adjustbox}{width=2\columnwidth}
    \def\arraystretch{1.1}
    \begin{tabular}{lcccccccccc}
        \toprule
        & \multicolumn{2}{c}{Initial} & \multicolumn{2}{c}{Fine-tune(small lr)} & \multicolumn{2}{c}{Fine-tune(large lr)} & \multicolumn{2}{c}{Pruning} & \multicolumn{2}{c}{Retrain after Pruning} \\
        \cmidrule(lr){2-3} \cmidrule(lr){4-5} \cmidrule(lr){6-7} \cmidrule(lr){8-9} \cmidrule(lr){10-11}
        & \textbf{main acc (\%)} & \textbf{wm acc (\%)} & \textbf{main acc (\%)} & \textbf{wm acc (\%)} & \textbf{main acc (\%)} & \textbf{wm acc (\%)} & \textbf{main acc (\%)} & \textbf{wm acc (\%)} & \textbf{main acc (\%)} & \textbf{wm acc (\%)} \\
        \midrule
        \texttt{WNN} & 95.84	& \textbf{100} & \textbf{97.77}	& \textbf{100}	& 96.55	& 86.36	& 12.87	& 0	& \underline{93.22}	& 0 \\
        \texttt{USP}  & \textbf{96.63} & \underline{99.64} & \underline{97.14} & \underline{96.32} & \textbf{96.86} & \underline{91.73} & \textbf{24.13} & \textbf{100} & 91.40 & \underline{41.96} \\
        \texttt{Margin} & \underline{96.60}  & 96.60   & 96.61  & \textbf{100}  & 91.43  & 0  & \underline{15.36}  & \underline{73.30} & 92.90 & 3.33  \\
        \texttt{Blind} & 93.90 & \textbf{100} & 96.69 & \textbf{100} & \underline{96.70} & \textbf{100} & 9.78 & 9.00 & 88.45 & 13.60 \\
        \textbf{DCT-AW}   & 96.23 & \textbf{100}   & 96.00  & \textbf{100}    & \textbf{96.86} & \textbf{100}    & 12.87   & \textbf{100} & \textbf{93.52}  & \textbf{93.69} \\
        \bottomrule
    \end{tabular}
    \end{adjustbox}
\end{table*}

\begin{table*}[!h]
    \centering
    \caption{Performance comparison on Fashion MNIST.}\label{table:dif_attack_fs_mnist}
    \begin{adjustbox}{width=2\columnwidth}
    \def\arraystretch{1.1}
    \begin{tabular}{lcccccccccc}
        \toprule
        & \multicolumn{2}{c}{Initial} & \multicolumn{2}{c}{Fine-tune(small lr)} & \multicolumn{2}{c}{Fine-tune(large lr)} & \multicolumn{2}{c}{Pruning} & \multicolumn{2}{c}{Retrain after Pruning} \\
        \cmidrule(lr){2-3} \cmidrule(lr){4-5} \cmidrule(lr){6-7} \cmidrule(lr){8-9} \cmidrule(lr){10-11}
        & \textbf{main acc (\%)} & \textbf{wm acc (\%)} & \textbf{main acc (\%)} & \textbf{wm acc (\%)} & \textbf{main acc (\%)} & \textbf{wm acc (\%)} & \textbf{main acc (\%)} & \textbf{wm acc (\%)} & \textbf{main acc (\%)} & \textbf{wm acc (\%)} \\
        \midrule
        \texttt{WNN}  & 84.51	& \textbf{100} & 87.43	& \textbf{100}	& 84.93	& 73.30	& \underline{11.20}	& \underline{26.20}	& 85.23	& 0 \\
        \texttt{USP}  & \underline{88.41} & \underline{99.93} & 89.17 & \underline{99.59} & 88.42 & \underline{97.40} & 10.38 & 0.00 & 85.04 & \underline{32.67} \\
        \texttt{Margin} & 85.64  & \textbf{100}   & \underline{92.63}  & \textbf{100}  & \textbf{96.92}  & 10  & 9.05  & 0.50 & \underline{89.79}  & 13.33  \\
        \texttt{Blind} & 82.25 & \textbf{100} & 87.33 & \textbf{100} & 85.23 & 63.3& 10.38 &  0.00 & 85.04 & \underline{32.67} \\
        \textbf{DCT-AW}  & \textbf{93.73} & \textbf{100}   & \textbf{95.73}  & \textbf{100}    & \underline{96.50} & \textbf{100}    & \textbf{13.96}  & \textbf{50.01} & \textbf{91.45}  & \textbf{50.15} \\
        \bottomrule
    \end{tabular}
    \end{adjustbox}
\end{table*}


\section{Experiment}
\label{c:experiment}


\subsection{Experimental Setting}

We conduct experiments on both classification and regression tasks. We apply several existing DNN watermarking methods to KAN as baselines, including \texttt{WNN}~\cite{adi2018turning}, \texttt{Margin}~\cite{kim2023margin}, \texttt{Blind}~\cite{li2019prove}, and \texttt{USP}~\cite{chien2024customized}. 
Dataset details are described in a later subsection. 
For evaluation metrics, we apply \textbf{main accuracy (main acc.)} and \textbf{watermark accuracy (wm acc.)} to evaluate the proposed approach and baselines. Specifically, main accuracy (main acc.) is measured by test accuracy (classification) or RMSE (regression). 
For, watermark accuracy (wm acc.), 
each of the testing instances can be used to verify the existence of the watermark in the model. We calculate the watermark detection rate as the number of testing instances successfully detected by the detector to the total number of testing instances. To facilitate reproducibility, we release all code, trained models, and supplementary materials at~\cite{rep_material}.

\subsection{Classification Task}
In this section, we demonstrate the effectiveness of our proposed method \texttt{DCT-AW} through experiments on a classification task using the MNIST~\cite{deng2012mnist} and Fashion MNIST datasets~\cite{xiao2017fashion}. Tables \ref{table:dif_attack_mnist} and \ref{table:dif_attack_fs_mnist} show the watermark embedding performance of different baselines and our method, including functionality preservation and robustness against removal attacks.

\subsubsection{Functionality Preservation}

The \emph{Initial} column of Table \ref{table:dif_attack_mnist} and Table \ref{table:dif_attack_fs_mnist} show the performance of the baseline methods and our proposed method \texttt{DCT-AW} in terms of functionality preservation. 
Our method achieves nearly identical performance in main accuracy (main acc.) compared to other baseline methods 
and achieving the best performance in Fashion MNIST. Furthermore, our method achieves the highest watermark accuracy (wm acc.). 
These results demonstrate that \texttt{DCT-AW} achieves a strong balance between functionality and watermark effectiveness.

\subsubsection{Fine-tuning Attack}

We fine-tune the watermark models for 8 epochs using small learning rates ($0.001$) and large learning rates ($0.01$). The results are shown in the \emph{Fine-tune(small lr)} and \emph{Fine-tune(large lr)} columns of Table \ref{table:dif_attack_mnist} and Table \ref{table:dif_attack_fs_mnist}. Fine-tuning with a small learning rate results in small changes to the model, thereby preserving the watermark information more effectively as all baseline's wm acc remain 100\% except \texttt{USP}. However, when using a larger learning rate for fine-tuning, the significant changes make it easier for the watermark information to be removed. Our proposed method demonstrated effective resistance to both learning rates, as it only achieves 100\% wm acc in both MNIST and Fashion MNIST datasets, showcasing its robustness.

\subsubsection{Pruning Attack}

We set the pruning rate to 60\%, as this already causes a significant reduction in the model's original performance. Pruning more weights would substantially degrade the model's original performance, making it unattractive to attackers. The experimental results are shown in the \emph{Pruning} column of Table \ref{table:dif_attack_mnist} and Table \ref{table:dif_attack_fs_mnist}. The results show that pruning effectively removes the watermark performance of many baselines across different datasets. Particularly on the Fashion MNIST dataset, the wm acc of \texttt{USP}, \texttt{Margin}, and \texttt{Blind} is significantly low, with this attack nearly completely compromising the watermark verification capability. In contrast, our proposed \texttt{DCT-AW} achieves relatively robust performance, demonstrating its resilience under pruning attacks. 

\subsubsection{Pruning with Retrain Attack}
We prune 60\% of the model parameters and then retrained the pruned model using a learning rate of $0.001$. The results are shown in the \emph{Retrain after Pruning} column of Table \ref{table:dif_attack_mnist} and Table \ref{table:dif_attack_fs_mnist}. It can be observed that after this attack, the wm acc of all baseline methods dropped significantly, falling below 50\% on the MNIST dataset and below 40\% on the Fashion MNIST dataset. 
However, our method, \texttt{DCT-AW}, embeds the watermark within the model's activation functions, effectively preserving the watermark information. 

\subsection{Regression Task}
In this section, we applied \texttt{DCT-AW} on regression task using Feynman dataset~\cite{udrescu2020ai, udrescu2020ai2}. The Feynman dataset collects many physics equations from Feynman's textbooks. It is noted that KAN can effectively model physical equations~\cite{liu2024kan}, making it a suitable choice for this experiment. 
We generate the dataset by randomly sampling $(u, v) \in (-1, 1)$ and compute corresponding output $f(u, v)$, then train a neural network to predict $f$ from $(u, v)$. 
We use RMSE as the main task metric to evaluate functionality preservation.

\begin{table*}[!t]
\centering
\caption{Experimental results with different Feynman equations.} 
\label{table:KAN_shapes}
\def\arraystretch{1.4}
\begin{tabular}{|c|c|c|c|c|c|c|}
\hline
\textbf{Formula} & \textbf{Variables} & \makecell{\Gape[2pt] {\textbf{Clean Model}} \\ \textbf{RMSE}} & \makecell{\Gape[2pt] {\textbf{Watermarked Model}} \\ \textbf{RMSE}} & \makecell{\Gape[2pt] {\textbf{Watermarked Model}} \\ \textbf{wm acc}} & \makecell{\Gape[2pt] {\textbf{Finetune}} \\ \textbf{wm acc}} & \makecell{\Gape[2pt] {\textbf{Retrain}} \\ \textbf{wm acc}}  \\ \hline
$\exp\left(-\frac{\theta^2}{2\sigma^2}\right) / \sqrt{2\pi\sigma^2}$ & $\theta, \sigma$ & 0.0653 & 0.0652 & 100 & 100 & 100 \\ \hline
$\exp\left(-\frac{(\theta - \theta_1)^2}{2\sigma^2}\right) / \sqrt{2\pi\sigma^2}$ & $\theta, \theta_1, \sigma$ & 1.0181 & 1.0181 & 100 & 100 & 100 \\ \hline
$\frac{a}{(b-1)^2 + (c-d)^2 + (e-f)^2}$ & $a, b, c, d, e, f$ & 2.0808 & 2.0809 & 100 & 100 & 50.00 \\ \hline
$1 + a \sin\theta$ & $a, \theta$ & 0.2946 & 0.2947 & 100 & 100 & 100 \\ \hline
$a \left(\frac{1}{b} - 1\right)$ & $a, b$ & 179.896 & 179.8957 & 100 & 94.35 & 94.35 \\ \hline
$\frac{1 - a}{\sqrt{1 - b^2}}$ & $a, b$ & 1.9631 & 1.964 & 100 & 100 & 100 \\ \hline
$\frac{a+b}{1+ab}$ & $a, b$ & 0.6718 & 0.6723 & 100 & 90.30 & 96.55 \\ \hline
$\frac{1+ab}{1+a}$ & $a, b$ & 11.6906 & 11.5592 & 100 & 100 & 100 \\ \hline
$\arcsin(n\sin\theta_2)$ & $n, \theta_2$ & 0.3106 & 0.3108 & 100 & 95.30 & 97.80 \\ \hline
$\frac{1}{1+ab}$ & $a, b$ & 1.7422 & 1.742 & 100 & 100 & 100 \\ \hline
$\sqrt{1 + a^2 - 2a\cos(\theta_1 - \theta_2)}$ & $a, \theta_1, \theta_2$ & 0.4188 & 0.4202 & 100 & 100 & 100 \\ \hline
$\frac{\sin^2\left(\frac{n\theta}{2}\right)}{\sin^2\left(\frac{\theta}{2}\right)}$ & $n, \theta$ & 0.2976 & 0.2983 & 100 & 100 & 100 \\ \hline
$n_0 e^{-a}$ & $n_0, a$ & 0.7451 & 0.7453 & 100 & 61.15 & 83.30 \\ \hline
$\cos a + a\cos^2a$ & $a, \alpha$ & 0.4041 & 0.4038 & 100 & 100 & 100 \\ \hline
$(a-1)b$ & $a, b$ & 0.6533 & 0.6539 & 100 & 97.25 & 100 \\ \hline
$\frac{1}{4\pi}c\sqrt{a^2 + b^2}$ & $a, b, c$ & 0.0364 & 0.0364 & 100 & 88.75 & 100 \\ \hline
$n_0(1 + a\cos\theta)$ & $n_0, a, \theta$ & 0.6474 & 0.6478 & 100 & 91.85 & 94.95 \\ \hline
$\frac{n \alpha}{1 - \frac{n \alpha}{3}}$ & $n_0, a$ & 0.3429 & 0.3429 & 100 & 93.10 & 97.70 \\ \hline
$\frac{n_0}{\exp(a)+\exp(-a)}$ & $n_0, a$ & 0.299 & 0.2991 & 100 & 89.75 & 89.75 \\ \hline
$a + \alpha b$ & $a, \alpha, b$ & 0.67 & 0.6702 & 100 & 98.85 & 98.85 \\ \hline
$\frac{a}{b}$ & $a, b$ & 179.9028 & 179.9031 & 100 & 95.85 & 95.85 \\ \hline
$a\frac{\sin^2\left(\frac{b-c}{2}\right)}{\left(\frac{b-c}{2}\right)^2}$ & $a, b, c$ & 0.2769 & 0.2771 & 100 & 98.00 & 98.00 \\ \hline
$\sqrt{1 + a^2 + b^2}$ & $a, b$ & 0.1603 & 0.1603 & 100 & 100 & 100 \\ \hline
$\beta(1 + \alpha\cos\theta)$ & $\alpha, \beta, \theta$ & 0.6474 & 0.6478 & 100 & 97.10 & 97.10 \\ \hline
\end{tabular}
\end{table*}

The experimental results are shown in Table~\ref{table:KAN_shapes}. By observing the RMSE of both clean models and watermark models, we find that in all experiments, the main task performance drops by less than 1\% after embedding the watermark. This indicates that our method has the capability for functionality preservation in regression tasks. We also examine the robustness of our watermarking method in regression tasks. The results indicate that after fine-tuning and retraining attacks, the wm acc remains above 90\% in most cases. 

\section{Conclusion}
\label{c:conclusion}

In this paper, we propose \emph{Discrete Cosine Transform-based Activation Watermarking (\texttt{DCT-AW})}, a watermarking method for Kolmogorov-Arnold Network (KAN), which perturbs activation functions using the discrete cosine transform (DCT) technique. This approach enables model owners to embed watermarks into any task that utilizes KAN as the training architecture. Experimental results demonstrate that our method outperform baselines and has a small impact on the original model's performance while effectively resisting various watermark removal attacks. 

\bibliographystyle{IEEEtran}
\bibliography{references}

\begin{thebibliography}{10}
\providecommand{\url}[1]{#1}
\csname url@samestyle\endcsname
\providecommand{\newblock}{\relax}
\providecommand{\bibinfo}[2]{#2}
\providecommand{\BIBentrySTDinterwordspacing}{\spaceskip=0pt\relax}
\providecommand{\BIBentryALTinterwordstretchfactor}{4}
\providecommand{\BIBentryALTinterwordspacing}{\spaceskip=\fontdimen2\font plus
\BIBentryALTinterwordstretchfactor\fontdimen3\font minus
  \fontdimen4\font\relax}
\providecommand{\BIBforeignlanguage}[2]{{%
\expandafter\ifx\csname l@#1\endcsname\relax
\typeout{** WARNING: IEEEtran.bst: No hyphenation pattern has been}%
\typeout{** loaded for the language `#1'. Using the pattern for}%
\typeout{** the default language instead.}%
\else
\language=\csname l@#1\endcsname
\fi
#2}}
\providecommand{\BIBdecl}{\relax}
\BIBdecl

\bibitem{li2019prove}
Z.~Li, C.~Hu, Y.~Zhang, and S.~Guo, ``How to prove your model belongs to you: A
  blind-watermark based framework to protect intellectual property of dnn,'' in
  \emph{Proceedings of the 35th annual computer security applications
  conference}, 2019, pp. 126--137.

\bibitem{kim2023margin}
B.~Kim, S.~Lee, S.~Lee, S.~Son, and S.~J. Hwang, ``Margin-based neural network
  watermarking,'' in \emph{International Conference on Machine Learning}.\hskip
  1em plus 0.5em minus 0.4em\relax PMLR, 2023, pp. 16\,696--16\,711.

\bibitem{chien2024customized}
T.-Y. Chien and C.-Y. Shen, ``Customized and robust deep neural network
  watermarking,'' in \emph{Proceedings of the 17th ACM International Conference
  on Web Search and Data Mining}, 2024, pp. 134--142.

\bibitem{liu2024kan}
Z.~Liu, Y.~Wang, S.~Vaidya, F.~Ruehle, J.~Halverson, M.~Soljacic, T.~Y. Hou,
  and M.~Tegmark, ``{KAN}: Kolmogorov{\textendash}arnold networks,'' in
  \emph{The Thirteenth International Conference on Learning Representations},
  2025.

\bibitem{kolmogorov1957representation}
A.~N. Kolmogorov, ``On the representation of continuous functions of many
  variables by superposition of continuous functions of one variable and
  addition,'' in \emph{Doklady Akademii Nauk}, vol. 114, no.~5.\hskip 1em plus
  0.5em minus 0.4em\relax Russian Academy of Sciences, 1957, pp. 953--956.

\bibitem{genet2024tkan}
R.~Genet and H.~Inzirillo, ``Tkan: Temporal kolmogorov-arnold networks,''
  \emph{arXiv preprint arXiv:2405.07344}, 2024.

\bibitem{cheon2024kolmogorov}
M.~Cheon, ``Kolmogorov-arnold network for satellite image classification in
  remote sensing,'' \emph{arXiv preprint arXiv:2406.00600}, 2024.

\bibitem{uchida2017embedding}
Y.~Uchida, Y.~Nagai, S.~Sakazawa, and S.~Satoh, ``Embedding watermarks into
  deep neural networks,'' in \emph{Proceedings of the 2017 ACM on international
  conference on multimedia retrieval}, 2017, pp. 269--277.

\bibitem{adi2018turning}
Y.~Adi, C.~Baum, M.~Cisse, B.~Pinkas, and J.~Keshet, ``Turning your weakness
  into a strength: Watermarking deep neural networks by backdooring,'' in
  \emph{27th USENIX security symposium (USENIX Security 18)}, 2018, pp.
  1615--1631.

\bibitem{zhang2018protecting}
J.~Zhang, Z.~Gu, J.~Jang, H.~Wu, M.~P. Stoecklin, H.~Huang, and I.~Molloy,
  ``Protecting intellectual property of deep neural networks with
  watermarking,'' in \emph{Proceedings of the 2018 on Asia Conference on
  Computer and Communications Security}, 2018, pp. 159--172.

\bibitem{fan2021deepipr}
L.~Fan, K.~W. Ng, C.~S. Chan, and Q.~Yang, ``Deepipr: Deep neural network
  ownership verification with passports,'' \emph{IEEE Transactions on Pattern
  Analysis and Machine Intelligence}, vol.~44, no.~10, pp. 6122--6139, 2021.

\bibitem{hornik1989multilayer}
K.~Hornik, M.~Stinchcombe, and H.~White, ``Multilayer feedforward networks are
  universal approximators,'' \emph{Neural networks}, vol.~2, no.~5, pp.
  359--366, 1989.

\bibitem{arnol1959representation}
V.~I. Arnol'd, ``On the representation of continuous functions of three
  variables by superpositions of continuous functions of two variables,''
  \emph{Matematicheskii Sbornik}, vol.~90, no.~1, pp. 3--74, 1959.

\bibitem{kolmogorov1961representation}
N.~Kolmogorov, \emph{On the representation of continuous functions of several
  variables by superpositions of continuous functions of a smaller number of
  variables}.\hskip 1em plus 0.5em minus 0.4em\relax American Mathematical
  Society, 1961.

\bibitem{abueidda2024deepokan}
D.~W. Abueidda, P.~Pantidis, and M.~E. Mobasher, ``Deepokan: Deep operator
  network based on kolmogorov arnold networks for mechanics problems,''
  \emph{arXiv preprint arXiv:2405.19143}, 2024.

\bibitem{ahmed1974discrete}
N.~Ahmed, T.~Natarajan, and K.~R. Rao, ``Discrete cosine transform,''
  \emph{IEEE transactions on Computers}, vol. 100, no.~1, pp. 90--93, 1974.

\bibitem{deng2012mnist}
L.~Deng, ``The mnist database of handwritten digit images for machine learning
  research,'' \emph{IEEE Signal Processing Magazine}, vol.~29, no.~6, pp.
  141--142, 2012.

\bibitem{udrescu2020ai}
S.-M. Udrescu and M.~Tegmark, ``Ai feynman: A physics-inspired method for
  symbolic regression,'' \emph{Science Advances}, vol.~6, no.~16, p. eaay2631,
  2020.

\bibitem{udrescu2020ai2}
S.-M. Udrescu, A.~Tan, J.~Feng, O.~Neto, T.~Wu, and M.~Tegmark, ``Ai feynman
  2.0: Pareto-optimal symbolic regression exploiting graph modularity,''
  \emph{Advances in Neural Information Processing Systems}, vol.~33, pp.
  4860--4871, 2020.

\bibitem{rep_material}
G.-J. Wu, C.-H. Lu, and C.-Y. Shen, ``Reproducibility materials (codes, models,
  and documents),'' in \emph{\url{https://reurl.cc/zqKLja}}, 2025.

\bibitem{xiao2017fashion}
H.~Xiao, K.~Rasul, and R.~Vollgraf, ``Fashion-mnist: a novel image dataset for
  benchmarking machine learning algorithms,'' \emph{arXiv preprint
  arXiv:1708.07747}, 2017.

\end{thebibliography}

\end{CJK}
\end{document}